# A GRASP algorithm for the Meal Delivery Routing Problem

Daniel Giraldo-Herrera [0009-0003-8243-723X] and David Álvarez-Martínez [0000-0001-8411-1936]

Department of Industrial Engineering, Universidad de Los Andes, Bogotá, Colombia
ds.giraldoh@uniandes.edu.co
d.alvarezm@uniandes.edu.co

**Abstract.** With the escalating demand for meal delivery services, this study delves into the Meal Delivery Routing Problem (MDRP) within the context of last-mile logistics. Focusing on the critical aspects of courier allocation and order fulfillment, we introduce a novel approach utilizing a GRASP metaheuristic. The algorithm optimizes the assignment of couriers to orders, considering dynamic factors such as courier availability, order demands, and geographical locations. Real-world instances from a Colombian delivery app form the basis of our computational analysis. Calibration of GRASP parameters reveals a delicate trade-off between solution quality and computational time. Comparative results with a simulation-optimization based study underscore GRASP's competitive performance, demonstrating strengths in fulfilling orders and routing efficiency across diverse instances. This research enhances operational efficiency in the burgeoning food delivery industry, shedding light on practical algorithms for last-mile logistics optimization.

**Keywords:** GRASP, Last-mile logistics, Meal Delivery Routing Problem (MDRP), Order fulfillment

## 1 Introduction

With the fast growth of the food delivery industry, mainly driven by the increasing demand for home delivery services, efficiency in meal delivery has become a critical factor for both businesses and consumers. Since the pandemic, meal delivery applications have taken on a solid position in everyday life worldwide, as meal delivery applications have become more visible to the average citizen and have established their position in the market. The companies operating as intermediaries between stores, restaurants, and end consumers reported significant growth during the pandemic; an example of this is the Domicilios.com application, which had an approximate growth of 50.6% between 2019 and 2020 [1]. Similarly, it is worth noting that the number of couriers working for these meal delivery applications increased significantly; an approximate increase of 62.5% was recorded between 2019 and 2020, reaching approximately 195,000 delivery drivers this last year in Colombia [2]. This problem manifests itself in a wide variety of contexts, from local restaurants to global meal delivery platforms, and its resolution has a direct impact on critical aspects such as operating cost, customer satisfaction, environmental sustainability, and stakeholders' quality of life.

To understand customer behavior on the use of meal delivery applications, it is necessary to understand how the business model of these applications works. These applications aim to provide customers with a home delivery service, having restaurants and home delivery companies as strategic allies [3]. The applications offers the user a broad portfolio of restaurants to which they can place orders, taking into account the geographical position of the user; in this way, the applications can be understood as a communication channel between the customer and the restaurants since they take the order from the customers, assign a courier who will deliver the order in the shortest possible time, and subsequently receive the payment from the customers, which is redistributed between the application, the restaurants and domiciliary [4]. Therefore, the condition for a restaurant to be available is based on its ability to offer food as it is demanded since the application must meet the condition of delivering orders to customers in the shortest possible time. The couriers' job is limited to picking up orders from restaurants and delivering them to customers on time and in optimal conditions. It is the job of the application to determine which orders should be delivered by each courier based on their geographic location and their type of vehicle [1, 2].

On the other hand, the stores linked to this type of application have a full-service contract for working with the application. Therefore, the application delivers and collects the customer's money. Based on this type of contract, this application charges between 8% and 27% on each sale. Each percentage that the application earns on the restaurants differs for each. The profits of the applications are not only centered on the interaction between courier and customer but also on the commission charged to the commercial partners. [5]. Finally, as far as users are concerned, this application is designed to provide customers with services (such as restaurants, pharmacies, and markets, among others) and to be able to place orders to be delivered in the shortest possible time [3].

Considering those mentioned above, the delivery business benefits the application, the couriers, and the restaurant since it generates high profits for every stakeholder [6]. These business models provide the couriers with labor flexibility by not having fixed working hours, but at the same time, this generates uncertainty regarding the monthly income of the courier. Additionally, the couriers are not protected against possible occupational accidents, especially considering that most of them are transported on bicycles or motorcycles, means of transportation that increase the risk of suffering an accident while working [7]. Generating satisfaction for all those involved is a latent challenge due to factors such as the fluctuating demand for orders, the number of couriers, the quality of the roads, and weather conditions, among others.

This challenge is found in last mile logistics, commonly referred to as the Meal Delivery Routing Problem (MDRP). It is defined as a problem in which an establishment dispatches an order requested by a customer using a delivery driver, ensuring that the food arrives in excellent condition as soon as possible [8]. Thus, the MDRP problem seeks to improve system planning in real-time courier platforms by considering the demand and the geographical locations of the points of interest (restaurants, couriers, customers), making adequate use of the courier's carrying and transportation capacity [9]. To implement a viable solution for MDRP, it is necessary to consider all kinds of variables in addition to the geographical locations of the actors, such as the number of couriers available for deliveries, order preparation time by restaurants, distances and travel times of the couriers that vary depending on the means of transport they use, the time windows to fulfill an order, among others.

This project presents an alternative solution to the MDRP by implementing a GRASP metaheuristic, an algorithm designed primarily to solve combinatorial problems with multiple solutions. To test the quality of solutions of the proposed algorithm, real-life test instances of one day of operations will be used. Next, an in-depth study of the MDRP, part of the set of problems called Dynamic Vehicle Routing Problem (DVRP), will be carried out.

The paper is structured as follows: Section 2 reviews related literature, Section 3 outlines the problem, Section 4 details the GRASP approach, Section 5 analyzes its performance, and Section 6 concludes and suggests future research directions.

## 2      Related work

The vehicle routing problem is one of the most studied topics in operations research. However, over time, these models have become increasingly complex in an attempt to mimic the complexities of real life. The Meal Delivery Routing Problem (MDRP) belongs to the Dynamic Vehicle Routing Problem (DVRP) class that incorporates pickups and deliveries. One of the most widely used variants for solving meal delivery problems today is the Same Day Delivery Routing Problem (SDDRP) [8 - 13].

As the name implies, the SDDRP problem refers to a situation in which a facility must prepare and ship an order placed by a customer that same day in a limited time window. The customer is unknown initially, so there is no prior route to fulfill the customer's request [8]. Within the SDDRP, the objective is to minimize the number of orders that cannot be fulfilled and, additionally, to answer questions such as how many vehicles will be needed to meet the demand efficiently and how to organize the routing of the vehicles effectively, among others [11, 12, 14]. At this juncture, a dynamic scenario is envisaged since, as the day progresses, new requests from customers for immediate order fulfillment are generated [15, 16]. Usually, it is solved with dynamic programming, which recurrently falls into the curse of dimensionality. The other approach that has been implemented is the Rolling Horizon, which is used to solve time-dependent models repeatedly, where, in each iteration, the planning interval advances [17, 18, 19, 20]

Some studies on MDRP consider that food preparation times and schedules are random and are, in turn, one of the main challenges of home food delivery; in these studies, they proposed a dynamic solution that offered greater efficiency in the collection and delivery of meals [8, 12, 13, 16, 17]. They developed a dynamic pick-up and delivery model using random variables using a methodology based on graph theory, mathematical programming, and simulations, considering logistic constraints, vehicle capacity, and time windows for pick-up and delivery [13, 20, 21]. On the other hand, studies identify that couriers working through these platforms do not have the freedom they are promised, directly affecting their welfare [14]. Make changes to the allocation algorithm so that the couriers working on these platforms have greater freedom over their work decisions and their welfare is prioritized [21, 22, 23, 24]

In [7] is highlighted as a computational framework that allows the evaluation of different MDRP solutions. This tool consists of integrated software built mainly in Python, with a database assembled with Docker and a simulator in SimPy that uses the optimizer and the OSRM (Open-Source Routing Machine) container. The tool in question consists of three modules: the first, in which in-stances and information that validate the simulator with the real world are load-ed; the second, in which the interactions of the actors in the model and the metrics to be estimated are defined; the third in which the services that given the information determine the outputs to be generated are established.

In [17, 25] are presented all the complications of the platform assignment operation, some of which are the order-grabbing conflicts between couriers, where multiple couriers are interested in one or some orders, but only one can perform the service. Each courier's delivery route obeys the constraints, including precedence, capacity, and time window constraints. The information of orders changes dynamically with the continuous input and output of customer orders. The positions of couriers change every second, which means that the status of riders is different at different decision moments. Solutions at previous steps will affect the solution space of the subsequent decision moments. They model the problem as a multi-step sequential decision-making process based on MDP. They propose an RL-based order recommendation method to generate order recommendation policies for different couriers.

A relevant factor in meal delivery is taking advantage of the couriers, maximizing their utilization, and focusing on the impact of the limited available resources. In [26] highlighted a computational approach in deep reinforcement learning; the overall objective is to maximize total profit by minimizing the expected delays, postponing, or declining orders with low/negative rewards, and assigning the orders to the most appropriate couriers. There are other solutions to this problem; [20,27] consider using fleet vehicles and drones to deliver. They modeled this problem as a sequential decision process, where decisions are made at specific points in time based on available information, utilizing a combination of deep reinforcement learning and heuristics to complement the routing for vehicles and drones. For drones, a first-in-first-out (FIFO) assignment approach is used, prioritizing drones idling at the depot. For vehicles, the heuristic assigns customers to vehicles based on insertion costs and feasibility.

## 3     Problem description

Users visit the delivery application throughout the day to place an order. Based on the user's address and active radius, the service application decides whether the user is in the service area and can place an order. If so, and if the user places an order, a delivery time is promised. Since all users within the service area can place an order and all such orders will be delivered, it may only be possible to meet the promised delivery time for some customers. To manage a good service and have a good utilization of addresses, the supplier dynamically adjusts the radius of the service area according to the expected and observed demand. The objective is determining feasible routes for couriers to complete the pick-up and delivery of orders within time windows, with the objective to optimize a single or multiple performance measures.

### 3.1     Users

Users play a pivotal role in meal delivery as consumers seek a convenient dining experience. Interacting with the dedicated app or platform, they browse diverse restaurant menus, customize orders to their preferences, provide delivery details, and track the real-time status of their meal. Through the app, users seamlessly engage in the entire ordering journey, from menu exploration to final payment and delivery confirmation. The user, as the final consumer, once purchased the product, has a time windows in which the user must receive the product so the user can give a positive rate of satisfaction and establish themselves as a repeat customer [6, 7, 8, 12, 13].

### 3.2     Apps/platform

The meal delivery app or platform is the linchpin in this dynamic system, connecting users, couriers, and stores. Its primary role is to offer users a user-friendly interface for browsing restaurant options, placing orders, and managing transactions. Beyond user interactions, the app is the communication hub, facilitating real-time updates and coordination between users, couriers, and stores. From order placement to payment, establishing time windows to pick up the orders at store and deliver the orders to users, processing and delivery tracking, the app streamlines the entire meal delivery experience [7, 8, 20, 24].

### 3.3     Couriers

Couriers serve as the physical link between users and restaurants, ensuring the smooth execution of meal deliveries. Their role begins with receiving order details through the app and picking up meals from designated stores. Couriers then navigate the delivery route to transport the food to the specified location. Their timely and accurate deliveries are crucial in providing users with a positive experience and maintaining the efficiency of the overall meal delivery system [6 - 8].

### 3.4    Restaurants

Stores, typically restaurants, are essential to the meal delivery network. Partnering with the delivery service, these establishments receive and process incoming orders through the app. Their role involves preparing meals promptly and making them available for courier pickup. By leveraging the meal delivery system, stores expand their reach and cater to a broader customer base without managing the complexities of delivery logistics, enhancing their overall accessibility and competitiveness in the market [12, 13, 16, 17, 18].

The MDRP problem can be divided into two parts; the first part consists of assigning a courier to an order based on different variables, such as the geographical positions of the store, the courier, and the customer. The second part refers to the layout of the route that the courier must follow to deliver the courier's address in the shortest possible time and the order in optimal condition, considering that each courier has a maximum capacity of orders that can be carried simultaneously. Thus, some metrics to be considered to measure the efficiency of the algorithm to be implemented as a solution to the present problem are the number of orders delivered, the average delivery time of an order, and the total order delivery time since a user requests the order [7, 9, 13].

The delivery process of an order is represented in Fig. 1, where the user purchases in his preferred restaurant, and these order details are sent to the courier assigned through the application, who will pick up the order at the restaurant specified for that order and then take it to the user.

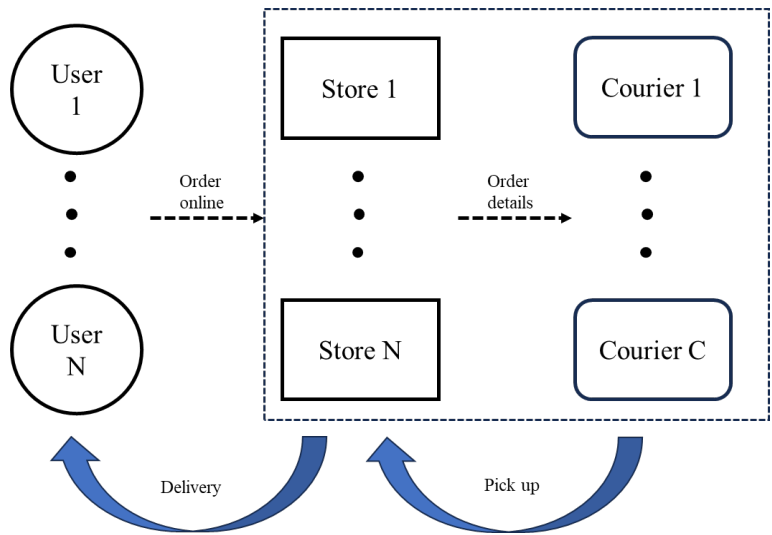

**Fig. 1.** Meal delivery process.

Let $U$ be the set of users, who want meals delivered and place orders, with each user $u \in U$ having a drop-off location $l_u$. Let $R$ be the set of restaurants, where meals must be picked up, and each restaurant $r \in R$ has a pick-up location $l_r$. Then, let $O$ be the set of orders placed by the users. Each order $o \in O$ has an associated restaurant $r_o \in R$ and user $u_o \in U$, a placement time $\alpha_o$, a ready time $e_o$, a pick-up service time $s^r$ (depending on the restaurant) and a mandatory drop-off service time $s^u$ linked to the user. Let $C$ be the set of couriers, who are used to deliver the orders in $O$. Let $c_o \in C$ be the courier assigned to order $o \in O$, noting that a courier may eventually not be assigned. Each courier $c \in C$ has an on-time $e_c$, an on-location $l_c$, and an off-time $l_c$, with $l_c > e_c$. A courier's compensation at the end of the shift is defined as $\sum_{m=1}^{i} p_i \, \forall \, r \in R$, where $p_i$ is the compensation for each order placed divided into a base amount and a variable amount for each order, which depends on the time it takes for an order to be accepted, and m is the number of orders placed. Orders from the same restaurant may be aggregated into bundles, or routes, where each route has a single pick-up but multiple drop-off locations. Let $S$ be the set of routes. Any route $s \in S$ must fulfill $|\{r_o, \forall \, o \in S\}| = 1$. The information of a user U associated with an order O is only disclosed to the couriers R at the time they agree to place the order. A courier can pick-up more than one order if they are coming from the same restaurant and the maximum capacity to carry is three orders. Fig. 2 is a general description of the logistics delivery.

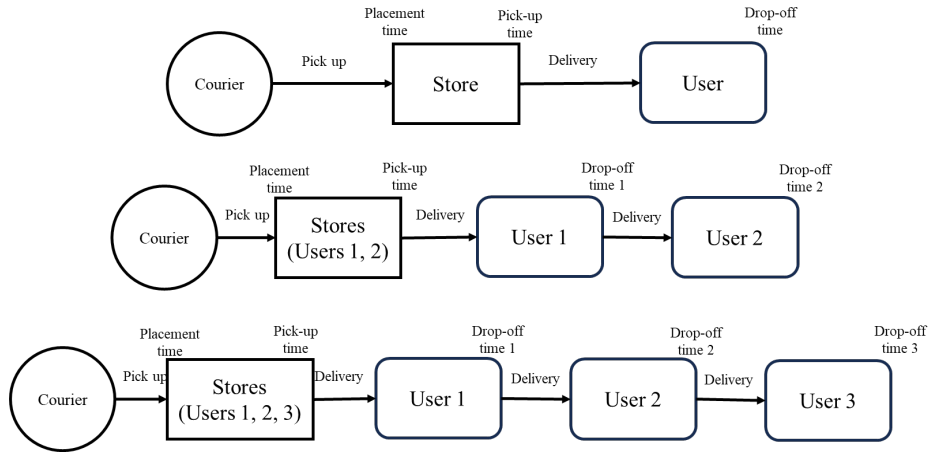

**Fig. 2.** Pick-up and delivery process considering time windows.

This detailed overview sets the stage for addressing the complexities inherent in the Meal Delivery Routing Problem, encompassing user interactions, dynamic service area adjustments, courier assignment, and optimizing delivery routes for enhanced operational efficiency. The subsequent sections will explore the metaheuristic approaches to address these challenges.

## 4    Methodology

To the best of the authors' knowledge, a GRASP-type metaheuristic algorithm tailored specifically for the Meal Delivery Routing Problem (MDRP) has yet to be introduced. The GRASP algorithm, pioneered by Feo and Resende [28], stands out as a highly adaptable tool for optimizing complex problems within real-world industrial settings [29]. Comprising two distinct stages—construction and improvement—GRASP's efficacy typically hinges on two primary parameters: *alpha* and *iterations*. *Alpha* governs the degree of randomness injected into decision-making when incorporating elements into solutions during the construction phase, while *iterations* denote the total count of both constructed and improved solutions. Below, we outline the essential components for aligning GRASP with MDRP.

**Algorithm 1.** Constructive_Phase

| |
|---|
| **Inputs:** couriers, orders; |
| **Parameters:** alpha |
| 1   assigned_orders ← Ø |
| 2   couriers_available ← **copy** couriers **all** |
| 3   orders_pending ← **copy** orders **all** |
| 4   **while** orders_pending **do** |
| 5       greedy_values ← Ø |
| 6       **for** courier_id, courier **in** couriers_available **do** |
| 7           **for** order_id, order **in** orders_pending **do** |
| 8               time travel ←_Calculate_distance_between_(courier, order)/courier_vel |
| 9               greedy_values ← greedy_values ∪ {time travel, courier_id, order_id} |
| 10          **end** |
| 11      **end** |
| 12      **sort** greedy_values **by** time travel |
| 13      RCL ← **copy** greedy_values alpha·_Size_(greedy_values) |
| 14      selected_courier_id, selected_order_id ← random from RCL |
| 15      **if** selected_courier_id **not in** assigned_orders **then** |
| 16          current_establishment ← orders[selected_order_id] [restaurant_ID'] |
| 17          assigned_orders[selected_courier_id] ← [current_establishment] |
| 18      **end** |
| 19      assigned_orders[selected_courier_id] ← selected_order_id |
| 20      **delete** selected_order_id **from** orders_pending |
| 21 **end** |
| 22 **return** assigned_orders |

The presented Algorithm 1, termed Constructive Phase, addresses the MDRP by iteratively assigning pending orders to available couriers. Initiated with an empty set to store assigned orders, copies of the original courier and order sets are maintained for iterative processing. The algorithm employs a greedy approach, calculating time travel between couriers and pending orders and sorting them to identify candidates depending on if the delivery time windows are feasible to be dropped off. A parameter, denoted as alpha, influences the selection of top candidates. Randomness is introduced in the selection process, ensuring flexibility in the assignment. If a selected courier has not been assigned any orders, the current establishment associated with the chosen order is recorded. The algorithm continues this process until all pending orders are assigned, efficiently allocating orders to couriers. This constructive phase, essential in initializing a solution for the MDRP, integrates deterministic and stochastic components, providing a foundation for subsequent optimization procedures.

**Algorithm 2.** Local_Search_Phase

| |
|---|
| **Inputs:** couriers, orders, assigned_orders |
| 1  better_assign ← **copy** assigned_orders **all** |
| 2  better_time_travel_total ← **for each assign in** assigned_orders **do** *Sum*(time travel) |
| 3  **for** courier_id_1, orders_assigned_1 **in** assigned_orders **do** |
| 4      **for** courier_id_2, orders_assigned_2 **in** assigned_orders **do** |
| 5        **if** courier_id_1 **is not** courier_id_2 **then** |
| 6            **for** order_id_1 **in** orders_assigned_1[1:] **do** |
| 7              **for** order_id_2 **in** orders_assigned_2[1:] **do** |
| 8                new_assign ← **copy** assigned_orders **all** |
| 9                **if** order_id_1 **in** new_assign[courier_id_1] **then** |
| 10                    **delete** order_id_1 **from** new_assign[courier_id_1] |
| 11                **end** |
| 12                **if** order_id_2 **in** new_assign [courier_id_2] **then** |
| 13                    **delete** order_id_2 **from** new_assign[courier_id_2] |
| 14                **end** |
| 15                new_assign [courier_id_1] ← [order_id_2] |
| 16                new_assign [courier_id_2] ← [order_id_1] |
| 17                    new_time_travel_total ← **for each assign in** new_assign **do** *Sum* (time_travel) |
| 18                **if** new_time_travel_total < better_time_travel_total **then** |
| 19                    better_assign ← **copy**(new_assign) |
| 20                    better_time_travel_total ← new_time_travel_total |
| 21                **end** |
| 22              **end** |
| 23            **end** |
| 24        **end** |
| 25    **end** |
| 26 **end** |
| 27 **return** better_assign |

Algorithm 2, denoted as Local Search Phase, is essential in refining solutions for the MDRP. The algorithm evaluates the total time travel based on the existing assignments in Algorithm 1. It then iterates through pairs of couriers and their assigned orders, considering potential swaps between orders. This iterative exploration involves creating a new assignment configuration by swapping selected orders between couriers, subsequently recalculating the total time traveled for each new configuration. If a resulting configuration yields a shorter total time travel, the assignment is updated to reflect this improvement. The algorithm continues this local search process until all possible swaps have been explored. The final output is an enhanced assignment of orders to couriers, effectively optimizing the total delivery time travel. This Local Search algorithm significantly contributes to refining and optimizing solutions obtained during the constructive phase, embodying an integral step in the iterative improvement process inherent in the MDRP resolution.

**Algorithm 3**. GRASP Algorithm

---

**Input:** couriers, orders;
**Parameters:** alpha, iterations

---

1  best_soltion $\leftarrow$ Ø
2  best_cost $\leftarrow \infty$
3  **for** i=1 **to** *iterations* **do**
4      constructed_assign $\leftarrow$ *Constructive_Phase*(couriers, orders; alpha)
5      improved_assign $\leftarrow$ *Local_Search_Phase*(couriers, orders, constructed_assign*)*
6      current_cost $\leftarrow$ **for each assign in** improved_assign **do** *Sum*(time_travel),
7      **if** current_cost < best_cost **then**
8          best_solution $\leftarrow$ improved_assign
9          best_cost $\leftarrow$ current_cost
10    **end**
11 **end**

---

In Algorithm 3 GRASP, initializing an empty set for the best solution and an infinite value for the best cost, the algorithm iteratively executes the GRASP Constructive Phase and subsequent Local Search. For each iteration, the total cost of the local solution is evaluated. If it outperforms the current best solution, the algorithm updates the best solution and cost. This process continues for the specified number of iterations. The algorithm concludes by returning this formatted solution, embodying a comprehensive approach to addressing the MDRP through GRASP, iterative refinement, and enhanced solution presentation.

Delivery assumptions: For consistency across iterations, we postulate that:

— Each order on the platform has a preparation time and time to be picked up and delivered, so this time window cannot be exceeded.
— There may be unfilled orders due to time windows, or not couriers near.
— If a courier arrives before the restaurant finishes the order, the courier should wait until the restaurant finishes the order.
— A courier can carry at most 3 orders in a single route if they are from the same restaurant.
— There are different vehicles for couriers, with different velocities; the fastest are the motorcycle with an average velocity of 20 km/h, the car with an average velocity of 15 km/h, the bicycle with an average velocity of 12 km/h, and walking average velocity is 5 km/h.
— When a courier delivers an order, he come back to the position and wait for another order to be delivered.

# 5    Computational experiments and result analysis

Our research addresses the shortfall in studies on the assignment of couriers allocated in different points of the city to deliver orders created by users at different restaurants, focusing on variables critical to operations optimization. The 22 instances used are based on real data from a delivery app in Colombia. Utilizing a Windows 11, 64-bit system with an Intel(R) Core (TM) i9-12900H CPU and 16 GB RAM, we developed an algorithm in Python for in-depth analysis.

## 5.1    Case of study

Each instance refers to a full day of orders and has data related to stores, couriers, and user orders. The information related to stores includes the store's ID and its location using latitude and longitude. In the information related to the couriers, there is the ID of each courier, the type of vehicle used to deliver the order, the initial location by latitude and longitude, the time it was connected to the order application, and the time it was disconnected. Finally, the information related to orders includes the order ID, the ID of the store to which the order is related, the user's location by latitude and longitude, the time the order was placed, the time the order preparation starts, the time the order is ready to be picked up to the courier and the time window the order must be delivered to the user's location.

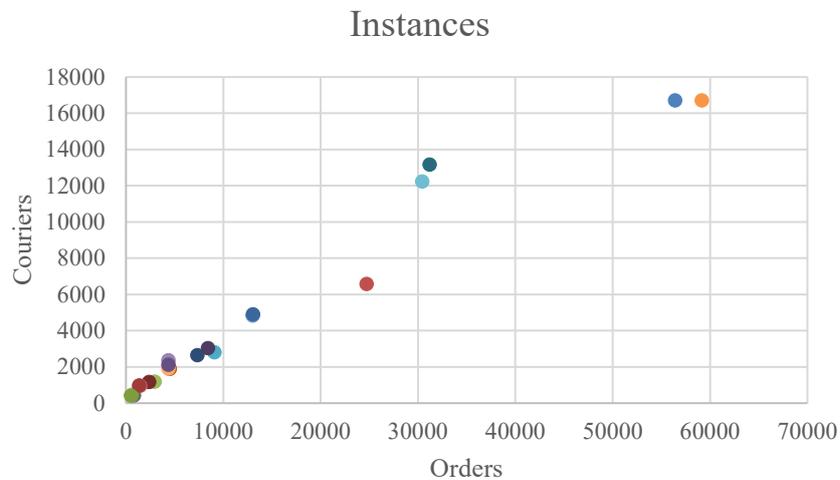

**Fig. 3.** Size instances for MDRP

**Fig. 3** shows data on instances, with each instance having associated values for the number of orders and dispatched couriers. We can draw conclusions from this data by analyzing the patterns and relationships between the number of orders and the dispatched couriers for each instance. Examining the ratio of couriers to the number of orders may provide insights into the efficiency of courier allocation. Instances with a lower ratio suggest effective courier utilization, while higher ratios indicate potential inefficiencies or challenges in meeting delivery demands.

## 5.2   Computational results

The platform's current policy is to run its allocation and routing algorithms every two minutes, with the objective of not having too many orders in queue and satisfying the customer's needs as soon as possible, which gives a higher satisfaction rate. With that in mind, we had to calibrate our GRASP algorithm to get results within that two-minute time window. First, we calibrate the alpha value with values between zero and one, increasing the value each time by 0.1. Second, we need to know the number of iterations for the algorithm execution; in this case, we started with 500 and increased iterations from 500 to 2000. The best results are in the alpha range of 0.7 and above and close to 2000 iterations. However, the computational time extended over five minutes, which is not a viable option from a business point of view since they must assign and route the couriers in less than two minutes to have a reasonable order fulfillment rate and generate good customer satisfaction. For this reason, we chose to use an alpha of 0.7 and 1000 iterations, which is 2% below the solutions with more iterations, but its computational time to obtain results is less than two minutes. **Fig. 4** shows the performance of the GRASP of the order fulfillment considering the number of iterations and the alpha value.

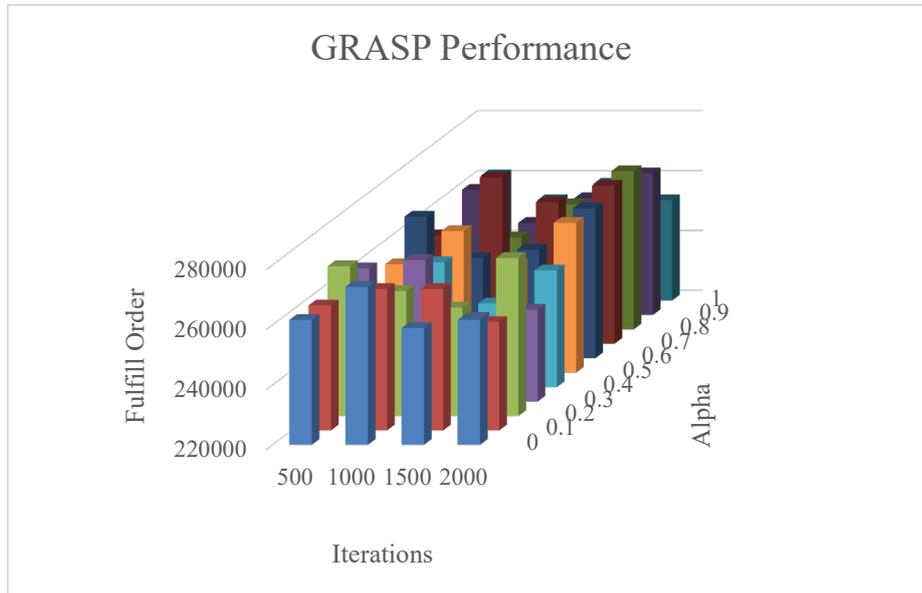

**Fig. 4.** Performance of the GRASP Algorithm, order fulfillment trade-off between number of iterations and alpha values

To check the efficiency of the GRASP algorithm, a comparison was made with a simulation study by [7]. He proposes an optimization simulation model for a food delivery company to test the efficiency of different optimization policies in a real-time geographic system. The corporate objective of these companies is to decrease the cost associated with transportation. However, their main objective is to complete as many orders as possible, so the primary comparison is the number of fulfilled orders. The platform these instances belong to uses its own map to estimate the routes. However, the simulator determines the routes through the haversine distance and the average speeds of each courier. Because of this, we have also implemented the exact measures of distances and average speeds of the couriers as the simulator in the GRASP algorithm for fair comparison purposes.

Table 1 shows the comparison between the two approaches; the first column is the index of the instance tested, the orders column specifies how many orders the instance has throughout the day, and the available courier's column refers to the available couriers throughout the day. CU means the number of couriers used for the solution of that instance. O.F. means the number of completed orders for that instance, and the routing time refers to the time it takes for the couriers to transport all completed orders. It was decided to work with the time rather than with the distance traveled because if a courier arrives before the order is ready in the restaurant, he must wait until it is finished to be able to take it to the user.

| C | Orders | Available Couriers | Sim-Opt [7] | | | GRASP | | | |
|---|---|---|---|---|---|---|---|---|---|
| | | | CU | O.F. | Routing time (min) | CU | O.F. | Routing time (min) | GAP (%) |
| 1 | 56420 | 16710 | 5548 | 54725 | 1676247 | **5572** | **55290** | 1699594 | **-1.02** |
| 2 | 24720 | 6570 | 2109 | 24114 | 721919 | **2108** | **24277** | 728141 | **-0.67** |
| 3 | 2959 | 1185 | 326 | 2878 | 86057 | **325** | **2879** | 86190 | **-0.03** |
| 4 | 844 | 415 | **91** | 821 | 25725 | 95 | **829** | 26110 | **-0.96** |
| 5 | 9115 | 2803 | 988 | 8879 | 295675 | **980** | **8965** | 300036 | **-0.95** |
| 6 | 59168 | 16708 | 5081 | 57347 | 1719146 | 5081 | 57347 | 1719179 | 0.00 |
| 7 | 7357 | 2643 | **887** | 7134 | 227203 | 893 | **7214** | 230046 | **-1.10** |
| 8 | 2407 | 1156 | **280** | 2336 | 67826 | 288 | **2351** | 68461 | **-0.63** |
| 9 | 735 | 455 | **97** | 718 | 21498 | 99 | **725** | 21945 | **-0.96** |
| 10 | 8451 | 3023 | **1190** | 8206 | 267422 | 1191 | **8287** | 270312 | **-0.97** |
| 11 | 31199 | 13152 | **3335** | 30223 | 857744 | 3357 | **30546** | 870390 | **-1.05** |
| 12 | 4516 | 1876 | **454** | 4394 | 128561 | 461 | **4419** | 129977 | **-0.56** |
| 13 | 13051 | 4824 | **1332** | 12708 | 366298 | 1338 | **12825** | 371105 | **-0.91** |
| 14 | 1462 | 941 | 146 | 1423 | 38806 | 146 | **1432** | 39122 | **-0.62** |
| 15 | 538 | 430 | **63** | 523 | 15361 | 66 | **527** | 15763 | **-0.75** |
| 16 | 4390 | 2368 | **423** | 4260 | 123298 | 436 | **4327** | 125834 | **-1.54** |

| 17 | 30448 | 12218 | **3266** | 29588 | 823150 | 3284 | **29884** | 833122 | **-0.99** |
| 18 | 4389 | 1908 | **438** | 4256 | 123848 | 440 | **4303** | 125578 | **-1.09** |
| 19 | 13076 | 4905 | 1303 | 12766 | 347941 | 1293 | **12863** | 351542 | **-0.75** |
| **20** | 1376 | 959 | 147 | 1341 | 37129 | 143 | **1355** | 37667 | **-1.03** |
| **21** | 539 | 396 | **68** | 524 | 14620 | 69 | **529** | 14951 | **-0.94** |
| **22** | 4387 | 2111 | **477** | 4230 | 123464 | 479 | **4285** | 125707 | **-1.28** |

**Table 1. Results obtained from 22 real-world instances.**

The comparison between the GRASP metaheuristic and the Optimization simulation across various instances reveals nuanced insights into their respective performances. Regarding the number of couriers used, the metaheuristic demonstrates competitive results, using more of them in the different instances, which is a good indicator due to the applications use this important resource. Analyzing the fulfillment of orders and routing time, the GRASP fulfills more orders than Optimization simulations, except in instance six where it delivers the same number of orders. Both algorithms showcase strengths in specific contexts, highlighting trade-offs between optimization objectives. Instances with varying order sizes and courier availability underscore the adaptability of the metaheuristic.

## 6     Conclusions

This study advances the field of last-mile logistics in the approach to food ordering applications by presenting an approximate approach through a GRASP algorithm adapted to optimize order fulfillment in the Meal Delivery Routing Problem.

GRASP improves the solutions obtained in previous models through various scenarios without losing sight of the needs of other stakeholders, such as making shipments with multiple couriers who will earn revenue for each delivery made and stores able to generate revenue through the applications.

The proposed methodology was tested under conditions reflective of the technological infrastructure of the Colombian company that inspired this study. The computational analyses demonstrate that our GRASP can seamlessly integrate into the application's computational framework, ensuring efficient operations without compromising user, courier, and restaurant response times, showing the potential of our approach to enhance the efficiency and effectiveness of meal delivery services within similar technological contexts.

As a future work, it is suggested to consider optimization models oriented to maximize the welfare of all the actors in the meal delivery services system. Seeking to guarantee a true democratization of the delivery service, given that this has become a vital service in our society, especially in large cities, and thuse be able to fulfill the sales promise of these services "connect a user who requires a product from a business with a delivery person willing to pick it up and deliver it" [30].